\documentclass[11pt]{article}

\usepackage{acl}
\usepackage{times}
\usepackage{latexsym}
\usepackage{comment}
\usepackage[T1]{fontenc}
\usepackage[utf8]{inputenc}
\usepackage[greek,english]{babel}
\usepackage{microtype}
\usepackage{inconsolata}
\usepackage{subcaption}
\usepackage{graphicx}
\usepackage{booktabs}
\usepackage{array}
\usepackage{hyperref}

\usepackage{lineno}

\title{Late Transformer Layers Recode Syntax Canonically: Evidence from
Greek Scrambling and Cross-Layer Generalisation}

\author{\normalfont%
\begin{tabular}{@{}>{\normalfont\centering\arraybackslash}p{0.48\textwidth}@{\hspace{0.02\textwidth}}>{\normalfont\centering\arraybackslash}p{0.48\textwidth}@{}}
  Christos Nikolaos Zacharopoulos & Revekka Kyriakoglou \\
  Independent Researcher & Universit\'e Paris 8 Vincennes--Saint-Denis \\
   & Paris, France \\
  \texttt{christonik@gmail.com} & \texttt{revekka.kyriakoglou@univ-paris8.fr} \\[8pt]
  Chara Tsoukala & Th\'eo Desbordes \\
  Institute for Language & Dept.\ of Basic Neurosciences \\
  and Speech Processing & Faculty of Medicine, University of Geneva \\
  Athena Research Center, Athens, Greece & Geneva, Switzerland \\
  \texttt{chara.tsoukala@athenarc.gr} & \texttt{theo.desbordes@unige.ch}
\end{tabular}%
}

\begin{document}
\maketitle

\begin{abstract}

Probing studies have established that syntactic information is decodable in
early and middle transformer layers, but what happens to that information in
later layers remains poorly understood. We apply a cross-layer
generalisation analysis to three Greek-tuned large language models evaluated
on tightly controlled minimal pairs: object-relative constructions in Modern
Greek, where canonical (Subject-Verb-Object; SVO) and non-canonical
(Verb-Subject-Object; VSO) orders differ only in within-clause word order,
while preserving propositional meaning. When a probe trained on late layers
(20--31) is tested on each early layer individually, it produces
below-chance transfer (cluster-corrected, $p<0.01$), classifying 99.3\% of
non-canonical sentences as canonical. Probe coefficients reverse sign around
layer~22, indicating a directional recoding toward the canonical form rather
than simple information loss. These findings characterise a representational
format change in late transformer layers that goes beyond the well-established
decline in syntactic decodability, and they generate a directly testable
prediction for human EEG and MEG decoding studies using the same stimuli.
Code and stimuli are publicly available on
\href{https://osf.io/5d3w8/overview?view_only=d42af279745543808cc377b9f96cb1af}{OSF}
\end{abstract}

\section{Introduction}\label{sec:introduction}

Consider two sentences in Modern Greek that carry the same meaning:

\vspace{6pt}
\noindent(1)\ \
\selectlanguage{greek}\textit{Η Άννα είδε τον σκύλο.}\selectlanguage{english}\hfill\textit{(canonical SVO)}\\
\hspace*{1.4em}\textsc{def.f.nom} Anna.\textsc{nom} see.\textsc{pst.3sg} \textsc{def.m.acc} dog.\textsc{acc}\\
\hspace*{1.4em}`Anna saw the dog.'

\vspace{4pt}
\noindent(2)\ \
\selectlanguage{greek}\textit{Είδε η Άννα τον σκύλο.}\selectlanguage{english}\hfill\textit{(non-canonical VSO)}\\
\hspace*{1.4em}see.\textsc{pst.3sg} \textsc{def.f.nom} Anna.\textsc{nom} \textsc{def.m.acc} dog.\textsc{acc}\\
\hspace*{1.4em}`Anna saw the dog.' \hfill lit. `Saw Anna the dog'
\vspace{6pt}

Because case endings mark grammatical roles, the verb may precede the subject
without altering the proposition \citep{georgiafentis2025information,
katsika2013processing}. A competent reader resolves both orders to the same
meaning. The question we address is not whether this resolution occurs, but
what representational changes enable it inside a transformer language model.

Probing studies since BERT have established that syntactic information peaks
in middle transformer layers and declines at the output.
\citet{tenney_bert_2019} showed that classical NLP tasks are resolved in a
layer-wise pipeline; \citet{tenney_what_2019} and \citet{hewitt_structural_2019}
confirmed that structural properties are most decodable in middle layers;
\citet{coenen_visualizing_2019} demonstrated geometric clustering of syntactic
relations in BERT's representation space
\citep[see also][]{belinkov_interpretability_2020}. These studies characterise
\textit{where} syntactic information resides, but not \textit{what happens to
it} after it peaks. Does the code simply weaken in later layers, or does the
representation change in a specific direction? Standard per-layer probing
cannot distinguish these possibilities, because it treats each layer
independently.

We address this gap with a cross-layer generalisation analysis borrowed from
temporal decoding in cognitive neuroscience \citep{king_characterizing_2014,
desbordes_temporal_2026}. A linear probe trained on representations at one
layer is tested on those at another; transfer failure between layers indicates
a change in representational format, and the direction of failure characterises
the nature of that change. This design has, to our knowledge, not previously been applied in
the probing literature on word-order representation in transformer LLMs.

Characterising these representational dynamics also matters for brain--model
alignment research, where layer depth correlates with cortical processing
stages \citep{caucheteux2021long, schrimpf_neural_2021,
goldstein_temporal_2025} but detailed comparisons reveal divergent mechanisms
in specific constructions \citep{zacharopoulos2023assessing,
ZACHAROPOULOS2026}.

Greek object-relative constructions provide a uniquely controlled test case: case
morphology marks grammatical roles independently of word order, so the same
words can appear in SVO or VSO order within a relative clause while preserving
propositional content, as in examples~(1)--(2). Syntactically rigid
languages do not permit this degree of experimental control.

We apply this analysis across all layers of three Greek-tuned transformer
models. A probe trained on layers~20--31 classifies 99.3\% of non-canonical
sentences as canonical when applied to early layers, producing below-chance
transfer across a contiguous cluster. Probe coefficients reverse sign around
layer~22, accounting for the directional inversion. This asymmetry generates a
directly testable prediction for human neural responses to the same stimuli.

\section{Materials \& Methods}\label{sec:methods}

\subsection{Models}

The primary model was \href{https://huggingface.co/ilsp/Llama-Krikri-8B-Base}{Llama-Krikri-8B-Base}
\citep{roussis2025krikri},
a Greek--English LLM based on the Llama-3.1-8B architecture with state-of-the-art
performance on Greek benchmarks. Two comparison variants were evaluated:
\href{https://huggingface.co/ilsp/Llama-Krikri-8B-Instruct}{Llama-Krikri-8B-Instruct}
(instruction-tuned) and
\href{https://huggingface.co/TheFinAI/plutus-8B-instruct}{Plutus-8B-Instruct}
(Low-Rank Adaptation on a Greek financial corpus).
Results for these variants are reported in Appendix~\ref{sec:appendix_gat_matrices}.

\subsection{Experimental Design and Stimuli}\label{sec:stimuli}

Stimuli were object-relative sentences in Modern Greek. The SVO/VSO distinction
refers to word order \textit{inside the embedded relative clause}, not the
matrix sentence; the matrix verb is always sentence-final in both conditions.
The SVO/VSO contrast from examples~(1)--(2) applies inside the embedded
relative clause:
[\textit{Det~N\textsubscript{2}~V\textsubscript{trans}}]
vs.\ [\textit{V\textsubscript{trans}~Det~N\textsubscript{2}}]; the
matrix subject and sentence-final verb remain fixed. Complete glossed
minimal-pair stimuli appear in
Appendix~\ref{sec:appendix_examples}.

Stimuli were systematically generated from a predefined lexicon of Greek
nouns, determiners, and verbs (25 masculine and 25 feminine human noun stems;
see Appendix~\ref{sec:appendix_lexicon_stimuli}). Template generation was
required because naturally occurring SVO/VSO pairs inevitably differ in
lexical content, whereas templates allow all lexical material to be held constant within each minimal pair.

Stem pairs for N\textsubscript{1} and N\textsubscript{2} were always distinct.
The design yielded $2^4 = 32$ unique fully counterbalanced linguistic
conditions. The main experiment used 128 sentences; a sensitivity analysis on
an expanded 1024-sentence set is reported in
Appendix~\ref{sec:appendix_gat_matrices}. All sentences were processed with the
model's native subword tokenizer (see Appendix~\ref{sec:tokenization}).

\subsection{Probing and Layer-wise Analysis}\label{sec:results1}

Word order information was framed as a binary classification problem (canonical
SVO vs.\ non-canonical VSO) over hidden-state summaries. For each sentence
and layer, the hidden-state matrix (tokens $\times$ hidden dimensions) for the
post-clause region was extracted. Four distributional statistics were computed
over the token dimension: mean, variance, skewness, and kurtosis. These four
scalars constitute the per-layer feature vector for a given sentence. This summary was chosen over raw hidden states because it provides a compact
characterisation of the sequence-level representational geometry and enables
the coefficient-sign analysis in \S\ref{sec:results}. Features
were normalised with a robust estimator (median and inter-quartile range). The
classifier was L2-regularised logistic regression with $C = 1.0$ (scikit-learn
default); no hyperparameter search was conducted, as the choice was fixed a
priori.

The clause boundary is the complementizer \foreignlanguage{greek}{που} (onset
of the relative clause); the pre-clause region contains all tokens up to and
including it, and the post-clause region, all tokens following it.

Layer-wise discriminative performance and cross-layer generalisation were
evaluated with ROC--AUC under stratified 10-fold cross-validation. To identify
contiguous ranges of layers with above- or below-chance performance, a
nonparametric, cluster-based one-sample permutation test was applied
(two-sided; 1000 permutations; cluster-forming threshold from the $t$
distribution; $\alpha = 0.01$) over AUC$-0.5$.

\subsection{Cross-layer Generalisation Analysis}\label{sec:cross_layer}

A single logistic-regression probe was trained on features pooled from
layers~20--31 and evaluated on each test layer~0--19. This GAT-style design
tests whether the late-layer representational format transfers to early
layers, rather than maximising per-layer accuracy.

\paragraph{Reproducibility.}
All stimuli, extracted activations, and analysis code are publicly available in
an anonymised OSF repository (code and stimuli under \texttt{code/}, activations
under \texttt{activations/}):
\url{https://osf.io/5d3w8/overview?view_only=d42af279745543808cc377b9f96cb1af}.

\section{Results}\label{sec:results}

\begin{figure*}[t]
    \centering
    \begin{subfigure}[b]{0.48\textwidth}
        \centering
        \includegraphics[height=3.6cm,width=\textwidth,keepaspectratio]%
          {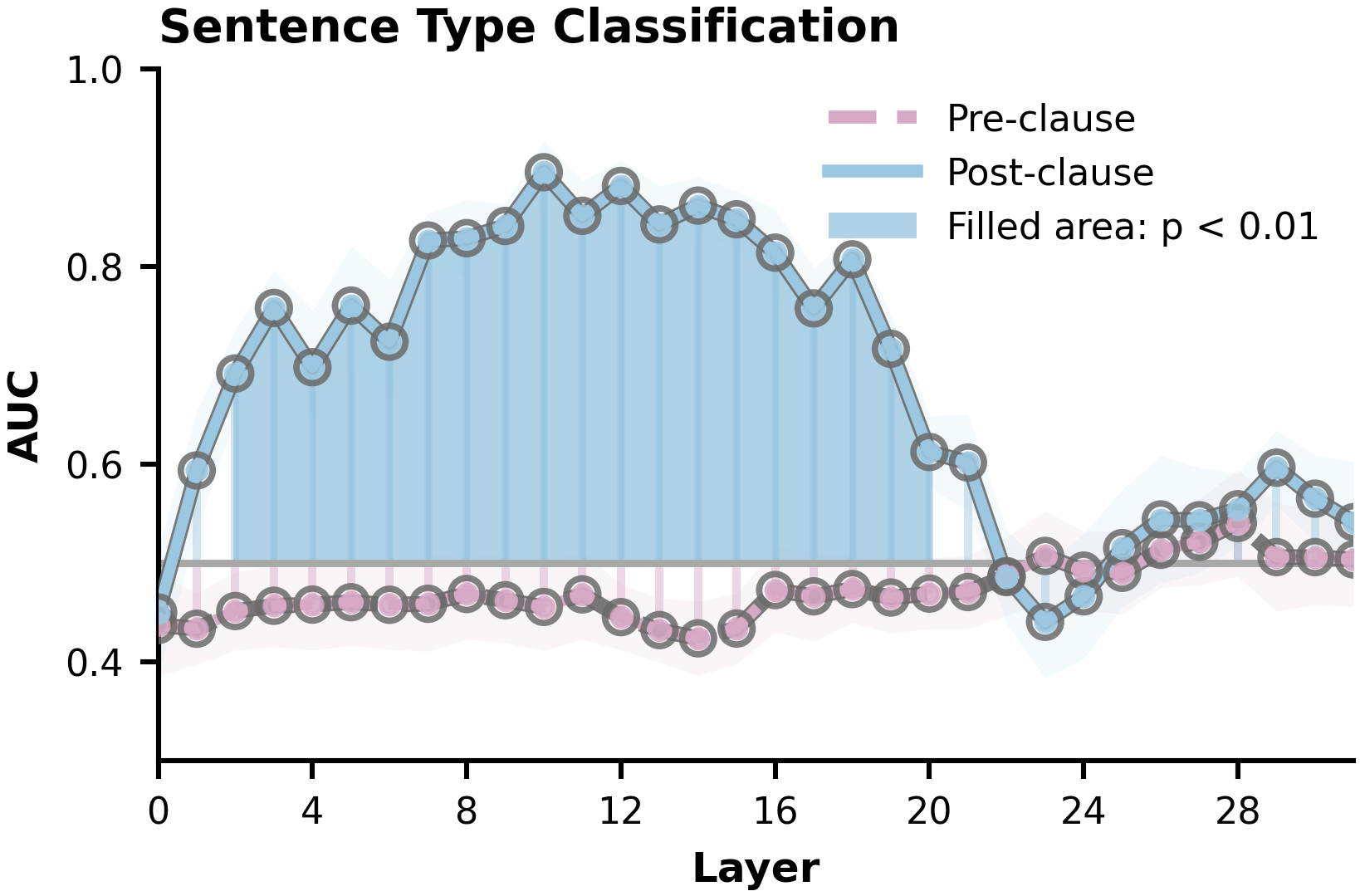}
        \caption{Layer-wise classification (canonical vs.\ non-canonical).
        Pre-clause (dashed) and post-clause (solid) ROC--AUC; shaded = SEM;
        bars = $p<0.01$ clusters.}
        \label{fig:auc_by_layer}
    \end{subfigure}
    \hfill
    \begin{subfigure}[b]{0.48\textwidth}
        \centering
        \includegraphics[height=3.6cm,width=\textwidth,keepaspectratio]%
          {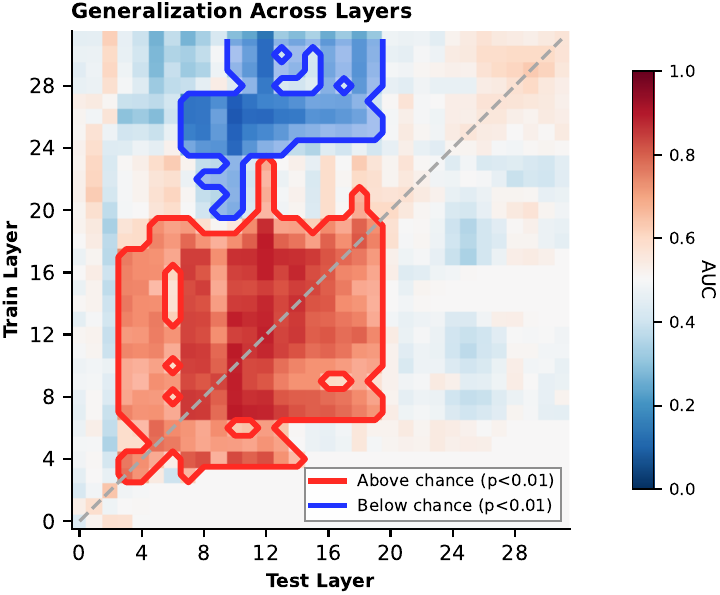}
        \caption{Cross-layer generalisation matrix. Each cell = ROC--AUC
        when a probe trained on one layer is tested on another. Contours
        mark $p<0.01$ clusters.}
        \label{fig:generalization_matrix}
    \end{subfigure}

    \vspace{4pt}

    \begin{subfigure}[b]{0.48\textwidth}
        \centering
        \includegraphics[height=3.6cm,width=\textwidth,keepaspectratio]%
          {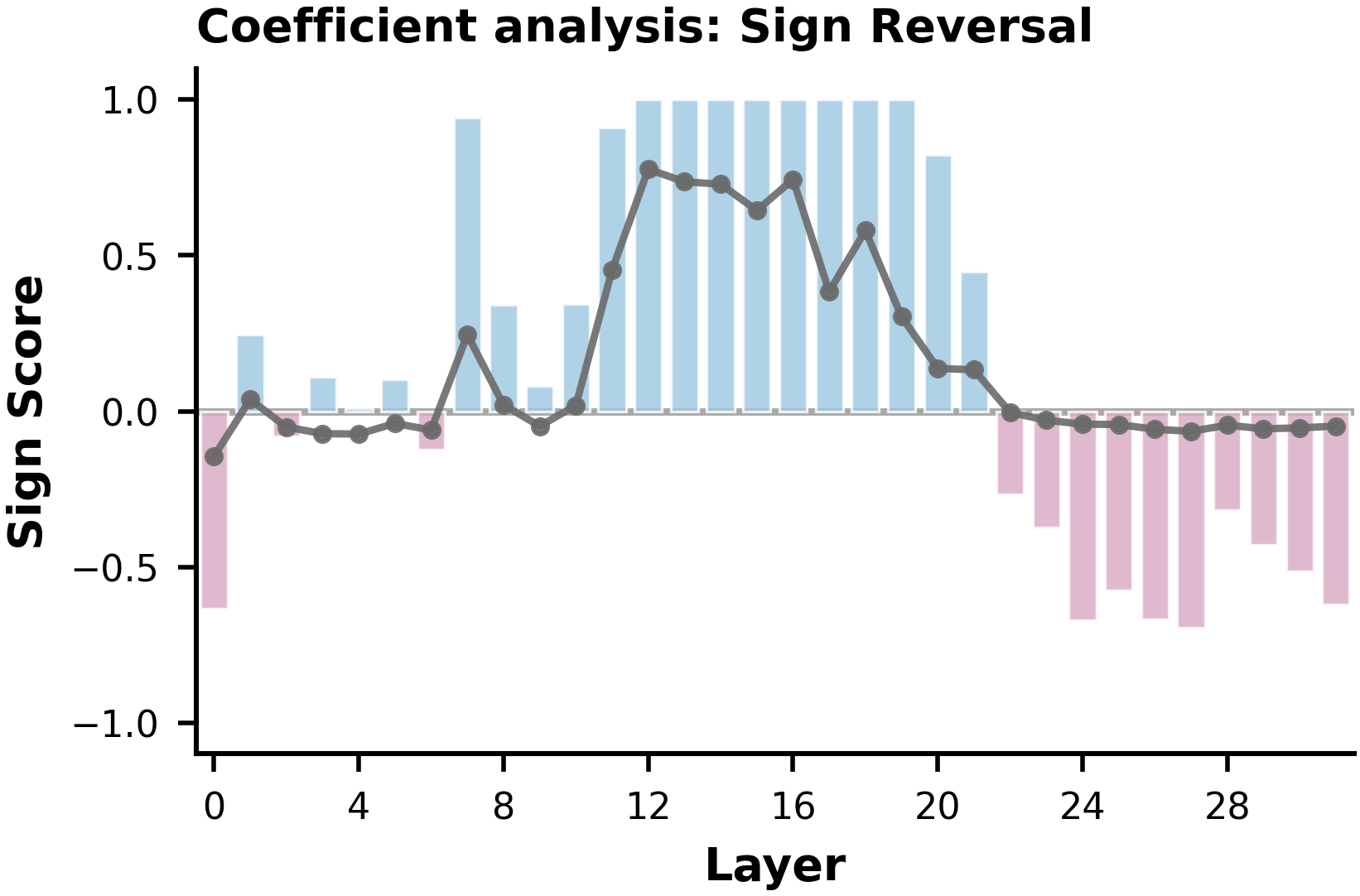}
        \caption{Probe coefficient sign score per layer. Positive = net
        positive coefficient mass; negative = net negative. Grey curve =
        mean normalised coefficient.}
        \label{fig:coefficients}
    \end{subfigure}
    \hfill
    \begin{subfigure}[b]{0.48\textwidth}
        \centering
        \includegraphics[height=3.6cm,width=\textwidth,keepaspectratio]%
          {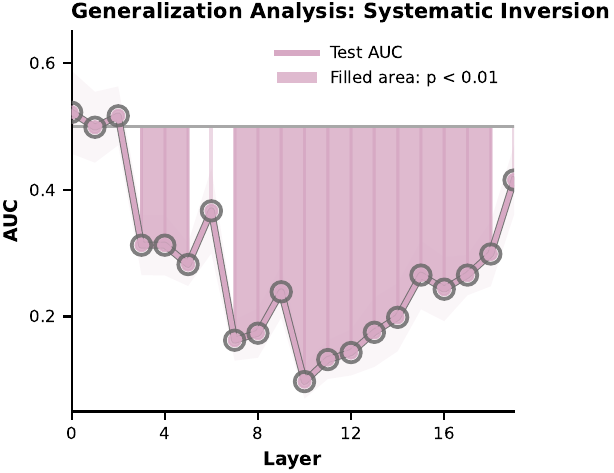}
        \caption{Late-to-early generalisation. AUC for a probe trained on
        layers~20--31, tested on each early layer. Shaded = significantly
        below chance ($p<0.01$).}
        \label{fig:generalization_analysis}
    \end{subfigure}
    \caption{Full results for Llama-Krikri-8B-Base (128-sentence set).
    (a)~Layer-wise AUC; (b)~cross-layer GAT matrix; (c)~coefficient sign
    dynamics; (d)~late-to-early transfer. The sign reversal around layer~22
    in~(c) is consistent with the directional below-chance transfer
    in~(d).}
    \label{fig:all_results}
\end{figure*}

\subsection{Sentence-type information emerges only after the clause boundary}

Post-clause activations yielded above-chance decoding performance from early
layers onward, peaking in the middle layers (Figure~\ref{fig:auc_by_layer}).
Pre-clause activations remained at chance level at all layers, as expected:
sentence-type information cannot be inferred before the structure-defining
constituents of the relative clause appear. Performance declined towards chance
in the final layers. The expanded 1024-sentence set and both comparison model
variants replicated the same qualitative layer-wise profile
(Appendix~\ref{sec:appendix_gat_matrices}).

\subsection{Middle layers support broad cross-layer generalisation}

Cross-layer generalisation analysis revealed a contiguous middle-layer region
(approximately layers~5--19) in which classifiers trained on one layer
generalised above chance to a wide range of test layers
(Figure~\ref{fig:generalization_matrix}).
This indicates a shared representational format for sentence type across this
portion of the network. The embeddings from late layers showed reduced cross-layer generalisation,
with AUC values approaching chance (0.50) by the final layers.

\subsection{Late-layer decision boundaries do not transfer to earlier layers}

A probe trained only on late-layer features (layers~20--31) produced
below-chance AUC when tested on earlier layers (Figure~\ref{fig:generalization_analysis}),
with significant effects across a contiguous cluster of early layers
($p < 0.01$). This result indicates that late-layer representations encode
sentence type in a different linear format from early-layer representations. On the significant early-layer cluster, the probe
classified 99.3\% of true non-canonical sentences as canonical and 98.0\% of
true canonical sentences as canonical, yielding an overall accuracy of 49.3\%.
The below-chance transfer therefore reflects a systematic directional bias:
the late-layer decision boundary, when applied to early-layer representations,
treats virtually all items as canonical regardless of their actual word order. The same qualitative late-to-early below-chance pattern was observed
in the 1024-sentence set and in both comparison models (Appendix~\ref{sec:appendix_lexicon_stimuli}).

\subsection{Probe coefficients reverse sign in later layers}

The directional nature of the transfer failure can be traced to the probe
weights themselves. We define a per-layer \textit{sign score} as the
proportion of positive probe coefficients minus the proportion of negative
ones; a score near $+1$ indicates that all coefficients are positive, a score
near $-1$ that all are negative. Around layer~22, the sign score crosses zero
and the dominant probe coefficients flip sign
(Figure~\ref{fig:coefficients}): the same distributional features that
predicted non-canonical class membership in early-to-middle layers predict
canonical class membership in late layers. This sign reversal is consistent
with a change in the direction of the linearly decodable contrast across
layers, and it explains why the late-layer boundary produces inverted,
rather than merely weak, transfer to early-layer representations.

\section{Discussion}\label{sec:discussion}

The central finding is that late transformer layers do not simply lose
syntactic sensitivity; they recode it in a specific direction. Prior probing
work established that syntactic information peaks in middle layers and declines
thereafter \citep{tenney_bert_2019, tenney_what_2019, hewitt_structural_2019,
coenen_visualizing_2019}. Those studies report \textit{where} information
resides at each layer; the present cross-layer analysis reveals \textit{how}
the representational format changes across layers. The decline is a directional
shift toward canonical-form representations, not a uniform weakening. This
converges with evidence that late layers support syntax-invariant operations
even in the semantic domain \citep{zacharopoulos2025machina}.

Three alternative accounts deserve consideration. Simple information loss
predicts near-chance transfer in both directions; it does not predict a 99.3\%
directional bias toward one class. Feature compression could produce biased
transfer, but the coordinated sign reversal across all four distributional
features around layer~22 is more consistent with a systematic representational
change than with incidental compression. Probe mismatch predicts noisy,
variable errors; instead, misclassification is nearly deterministic. We
therefore interpret the evidence as most consistent with directional recoding,
while acknowledging that causal intervention (e.g., ablation of late-layer
representations) would be needed to establish this conclusively.

Because SVO order and canonical status are co-extensive in this design, the
recoding could reflect a specifically syntactic shift or a more general
compression toward the high-frequency form (see Limitations).

Since transformers allow direct inspection of internal states, the
cross-layer analysis generates a testable prediction for human neural data:
the GAT-style design could be replicated with EEG or MEG recordings from
participants exposed to the same Greek stimuli, testing whether temporal
generalisation over neural responses shows a corresponding directional
asymmetry. Cross-linguistic extension and constructions where canonicality and
surface word order dissociate remain important directions for future work.

\section{Conclusion}\label{sec:conclusion}

Controlled scrambling in Greek allowed us to isolate syntactic structure from
semantic content and track how word-order representations evolve across
transformer layers. Syntactic information peaks in a mid-layer zone of stable
representations and is recoded in late layers into a format directionally
aligned with the canonical word order (99.3\% non-canonical$\to$canonical
misclassification; sign reversal at layer~22). These findings characterise a
directional recoding process whose signature is directly testable in human
EEG and MEG data.

\section{Limitations}

Four constraints qualify our interpretation. First, linear probes detect only
linearly decodable information; nonlinear representations may persist in late
layers. Second, SVO order and canonical status are co-extensive in this
design, so the probe may track either dimension. Third, residual lexical
regularities in template-generated stimuli might contribute weakly to
classification, although the balanced design minimises this; Jabberwocky
variants could isolate structural form more fully. Fourth, the study examines
a single language and construction type; cross-linguistic and
cross-constructional generalisation remains open.

\section{Ethical Considerations}

This study uses synthetic stimuli and pretrained language models. No human
participants were recruited and no personal data were collected. We identify
no direct participant-related ethical risks.

\bibliography{custom}

@article{georgiafentis2025information,
  title={Information structure in Greek: Interface and comparative studies},
  author={Georgiafentis, Michalis and Skopeteas, Stavros and Tsokoglou, Angeliki},
  journal={Journal of Greek Linguistics},
  volume={25},
  number={1},
  pages={3--9},
  year={2025},
  publisher={Brill}
}

@inproceedings{katsika2013processing,
  title={Processing subject and object relative clauses in a flexible word order language: Evidence from Greek},
  author={Katsika, Kalliopi and Allen, Shanley},
  booktitle={2013 Conference on Architectures and Mechanisms in Language Processing, Marseille, France},
  year={2013}
}

@inproceedings{zacharopoulos2023assessing,
  title={Assessing the influence of attractor-verb distance on grammatical agreement in humans and language models},
  author={Zacharopoulos, Christos and Desbordes, Th{\'e}o and Sabl{\'e}-Meyer, Mathias},
  booktitle={Proceedings of the 2023 Conference on Empirical Methods in Natural Language Processing},
  pages={16081--16090},
  year={2023}
}

@article{goldstein_temporal_2025,
	title = {Temporal structure of natural language processing in the human brain corresponds to layered hierarchy of large language models},
	volume = {16},
	copyright = {2025 The Author(s)},
	issn = {2041-1723},
	url = {https://www.nature.com/articles/s41467-025-65518-0},
	doi = {10.1038/s41467-025-65518-0},
	language = {en},
	number = {1},
	urldate = {2025-12-01},
	journal = {Nature Communications},
	publisher = {Nature Publishing Group},
	author = {Goldstein, Ariel and Ham, Eric and Schain, Mariano and Nastase, Samuel A. and Aubrey, Bobbi and Zada, Zaid and Grinstein-Dabush, Avigail and Gazula, Harshvardhan and Feder, Amir and Doyle, Werner and Devore, Sasha and Dugan, Patricia and Friedman, Daniel and Brenner, Michael and Hassidim, Avinatan and Matias, Yossi and Devinsky, Orrin and Siegelman, Noam and Flinker, Adeen and Levy, Omer and Reichart, Roi and Hasson, Uri},
	month = nov,
	year = {2025},
	pages = {10529},
}

@inproceedings{hewitt_structural_2019,
	address = {Minneapolis, Minnesota},
	title = {A {Structural} {Probe} for {Finding} {Syntax} in {Word} {Representations}},
	url = {https://aclanthology.org/N19-1419/},
	doi = {10.18653/v1/N19-1419},
	urldate = {2025-06-09},
	booktitle = {Proceedings of the 2019 {Conference} of the {North} {American} {Chapter} of the {Association} for {Computational} {Linguistics}: {Human} {Language} {Technologies}, {Volume} 1 ({Long} and {Short} {Papers})},
	publisher = {Association for Computational Linguistics},
	author = {Hewitt, John and Manning, Christopher D.},
	editor = {Burstein, Jill and Doran, Christy and Solorio, Thamar},
	month = jun,
	year = {2019},
	pages = {4129--4138},
}

@article{schrimpf_neural_2021,
	title = {The neural architecture of language: {Integrative} modeling converges on predictive processing},
	volume = {118},
	shorttitle = {The neural architecture of language},
	url = {https://www.pnas.org/doi/full/10.1073/pnas.2105646118},
	doi = {10.1073/pnas.2105646118},
	number = {45},
	urldate = {2022-05-24},
	journal = {Proceedings of the National Academy of Sciences},
	publisher = {Proceedings of the National Academy of Sciences},
	author = {Schrimpf, Martin and Blank, Idan Asher and Tuckute, Greta and Kauf, Carina and Hosseini, Eghbal A. and Kanwisher, Nancy and Tenenbaum, Joshua B. and Fedorenko, Evelina},
	month = nov,
	year = {2021},
	pages = {e2105646118},
}

@article{desbordes_temporal_2026,
	title = {Temporal evolution of neural codes: {The} added value of a geometric approach to linear coefficients},
	volume = {327},
	issn = {1053-8119},
	shorttitle = {Temporal evolution of neural codes},
	url = {https://www.sciencedirect.com/science/article/pii/S1053811926000558},
	doi = {10.1016/j.neuroimage.2026.121737},
	urldate = {2026-01-27},
	journal = {NeuroImage},
	author = {Desbordes, Théo and Olasagasti, Itsaso and Piron, Nicolas and Schwartz, Sophie and Kazanina, Nina},
	month = feb,
	year = {2026},
	pages = {121737},
}

@inproceedings{tenney_bert_2019,
	title = {{BERT} {Rediscovers} the {Classical} {NLP} {Pipeline}},
	booktitle = {Proceedings of the 57th {Annual} {Meeting} of the {Association} for {Computational} {Linguistics}},
	author = {Tenney, Ian and Das, Dipanjan and Pavlick, Ellie},
	year = {2019},
	pages = {4593--4601},
}

@inproceedings{tenney_what_2019,
	title = {What Do You Learn from Context? {Probing} for Sentence Structure in Contextualized Word Representations},
	booktitle = {Proceedings of the 7th {International} {Conference} on {Learning} {Representations}},
	author = {Tenney, Ian and Xia, Patrick and Chen, Berlin and Wang, Alex and Poliak, Adam and McCoy, R. Thomas and Kim, Najoung and Van Durme, Benjamin and Bowman, Samuel R. and Das, Dipanjan and Pavlick, Ellie},
	year = {2019},
}

@article{coenen_visualizing_2019,
	title = {Visualizing and Measuring the Geometry of {BERT}},
	journal = {Advances in Neural Information Processing Systems},
	volume = {32},
	author = {Coenen, Andy and Reif, Emily and Kim, Been and Pearce, Adam and Vi{\'e}gas, Fernanda and Wattenberg, Martin},
	year = {2019},
}

@inproceedings{belinkov_interpretability_2020,
	address = {Online},
	title = {Interpretability and {Analysis} in {Neural} {NLP}},
	url = {https://aclanthology.org/2020.acl-tutorials.1/},
	doi = {10.18653/v1/2020.acl-tutorials.1},
	urldate = {2026-03-13},
	booktitle = {Proceedings of the 58th {Annual} {Meeting} of the {Association} for {Computational} {Linguistics}: {Tutorial} {Abstracts}},
	publisher = {Association for Computational Linguistics},
	author = {Belinkov, Yonatan and Gehrmann, Sebastian and Pavlick, Ellie},
	editor = {Savary, Agata and Zhang, Yue},
	month = jul,
	year = {2020},
	pages = {1--5},
}

@article{king_characterizing_2014,
	title = {Characterizing the dynamics of mental representations: the temporal generalization method},
	volume = {18},
	issn = {1364-6613, 1879-307X},
	shorttitle = {Characterizing the dynamics of mental representations},
	url = {https://www.cell.com/trends/cognitive-sciences/abstract/S1364-6613(14)00019-9},
	doi = {10.1016/j.tics.2014.01.002},
	language = {English},
	number = {4},
	urldate = {2025-06-16},
	journal = {Trends in Cognitive Sciences},
	publisher = {Elsevier},
	author = {King, Jean-R{\'e}mi and Dehaene, Stanislas},
	month = apr,
	year = {2014},
	pages = {203--210},
}

@article{caucheteux2021long,
  title={Long-range and hierarchical language predictions in brains and algorithms},
  author={Caucheteux, Charlotte and Gramfort, Alexandre and King, Jean-Remi},
  journal={arXiv preprint arXiv:2111.14232},
  year={2021}
}

@article{roussis2025krikri,
  title={Krikri: Advancing Open Large Language Models for Greek},
  author={Roussis, Dimitris and Voukoutis, Leon and Paraskevopoulos, Georgios and Sofianopoulos, Sokratis and Prokopidis, Prokopis and Papavasileiou, Vassilis and Katsamanis, Athanasios and Piperidis, Stelios and Katsouros, Vassilis},
  journal={Findings of the Association for Computational Linguistics: EMNLP 2025},
  pages={5012–5033},
  year={2025}
}

@article{ZACHAROPOULOS2026,
title = {Disentangling Hierarchical and Sequential Computations during Sentence Processing},
journal = {Cortex},
year = {2026},
issn = {0010-9452},
doi = {https://doi.org/10.1016/j.cortex.2026.02.004},
url = {https://www.sciencedirect.com/science/article/pii/S0010945226000456},
author = {Christos-Nikolaos Zacharopoulos and Stanislas Dehaene and Yair Lakretz}
}

@inproceedings{zacharopoulos2025machina,
  title={In Machina {N400}: Pinpointing Where a Causal Language Model Detects Semantic Violations},
  author={Zacharopoulos, Christos-Nikolaos and Kyriakoglou, Revekka},
  booktitle={Artificial Intelligence and Cognitive Science},
  series={Communications in Computer and Information Science},
  volume={2950},
  publisher={Springer},
  year={2025}
}

\clearpage
\appendix

\section{Cross-model GAT matrices}
\label{sec:appendix_gat_matrices}

This appendix reports the generalization-across-layers (GAT) matrices for the
two comparison models emphasized in the manuscript, both evaluated on the
expanded $1024$-sentence stimulus set rather than the main $128$-sentence set. In
both cases, we observe the same qualitative structure described in the main
text: a broad mid-layer regime of above-chance generalization together with a
late-to-early transfer regime that falls below chance. The precise extent of
the clusters varies by model, but the overall representational organization is
preserved, indicating that the main findings replicate across comparison
systems even under the larger stimulus regime.

\begin{figure*}[t]
    \centering
    \begin{subfigure}[t]{0.48\textwidth}
        \centering
        \includegraphics[width=\textwidth]{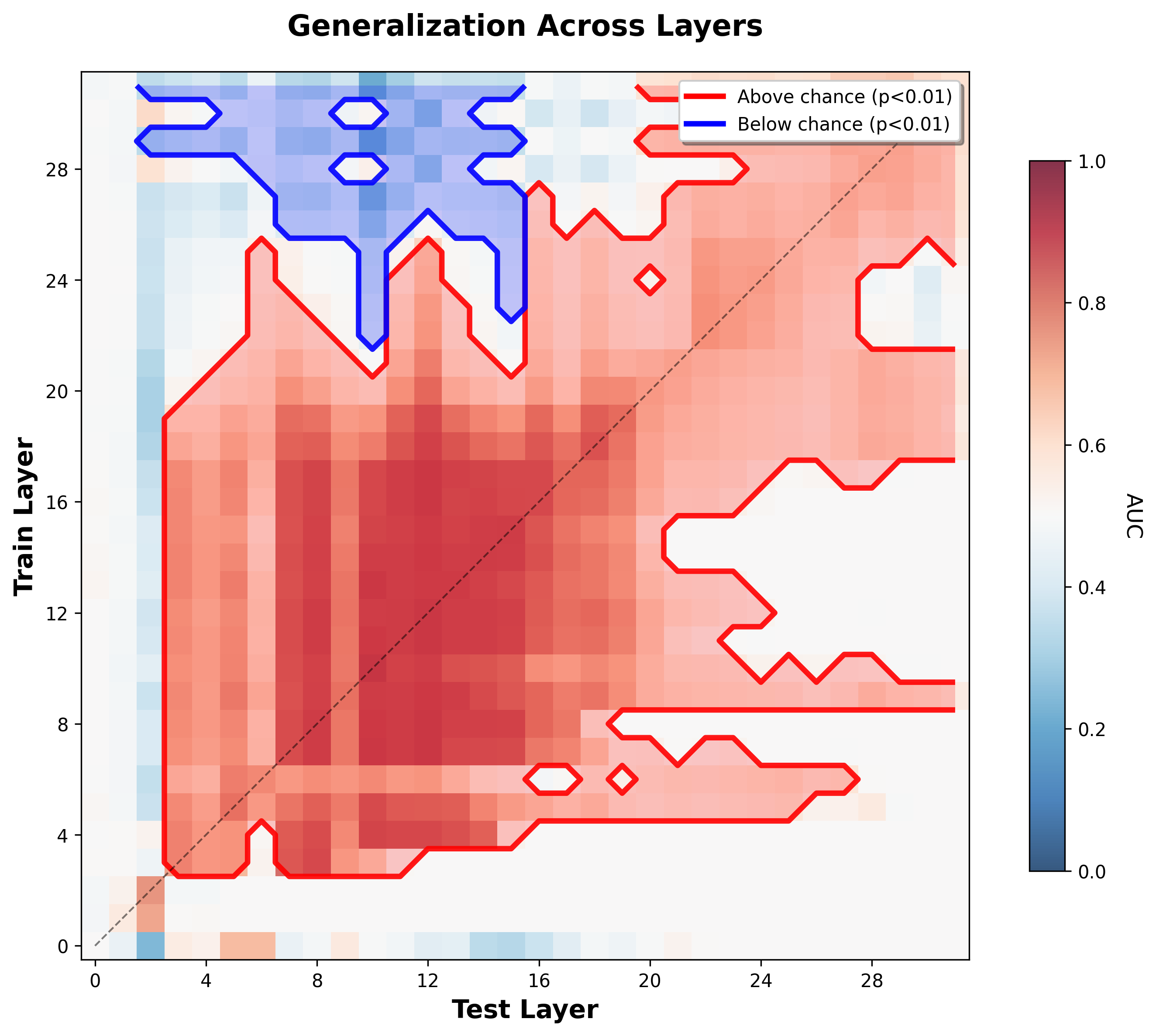}
        \caption{KriKri-Instruct}
        \label{fig:appendix_gat_krikri}
    \end{subfigure}
    \hfill
    \begin{subfigure}[t]{0.48\textwidth}
        \centering
        \includegraphics[width=\textwidth]{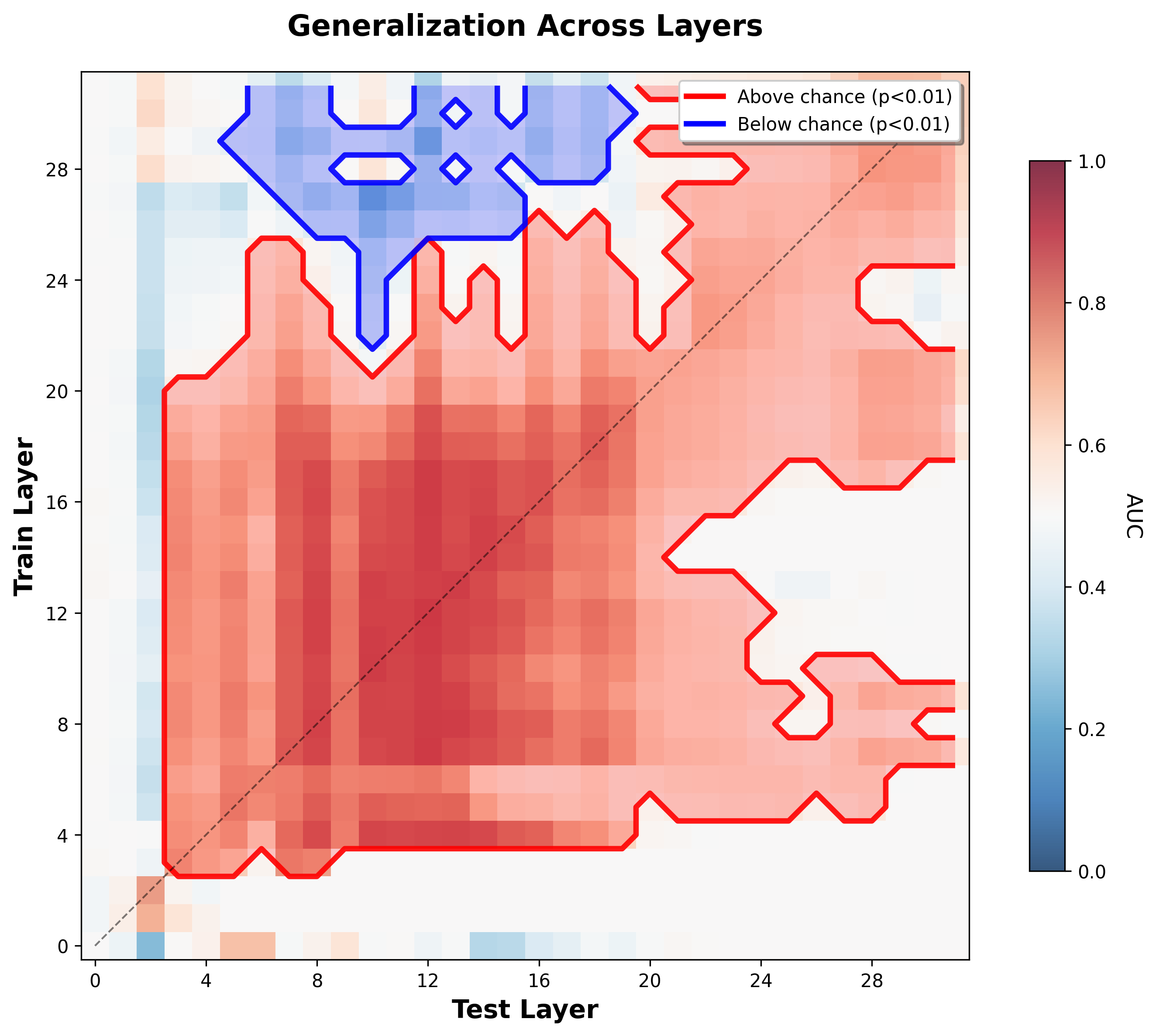}
        \caption{Plutus}
        \label{fig:appendix_gat_plutus}
    \end{subfigure}
    \caption{Generalization-across-layers matrices for KriKri-Instruct and
    Plutus, both evaluated on the expanded $1024$-sentence stimulus set. In
    both models, the dominant pattern is preserved: a broad zone of
    above-chance generalization in middle layers and a later regime showing
    below-chance transfer to earlier layers.}
    \label{fig:appendix_gat_models}
\end{figure*}

\begin{figure*}[t]
    \centering
    \begin{subfigure}[t]{0.48\textwidth}
        \centering
        \includegraphics[width=\textwidth]{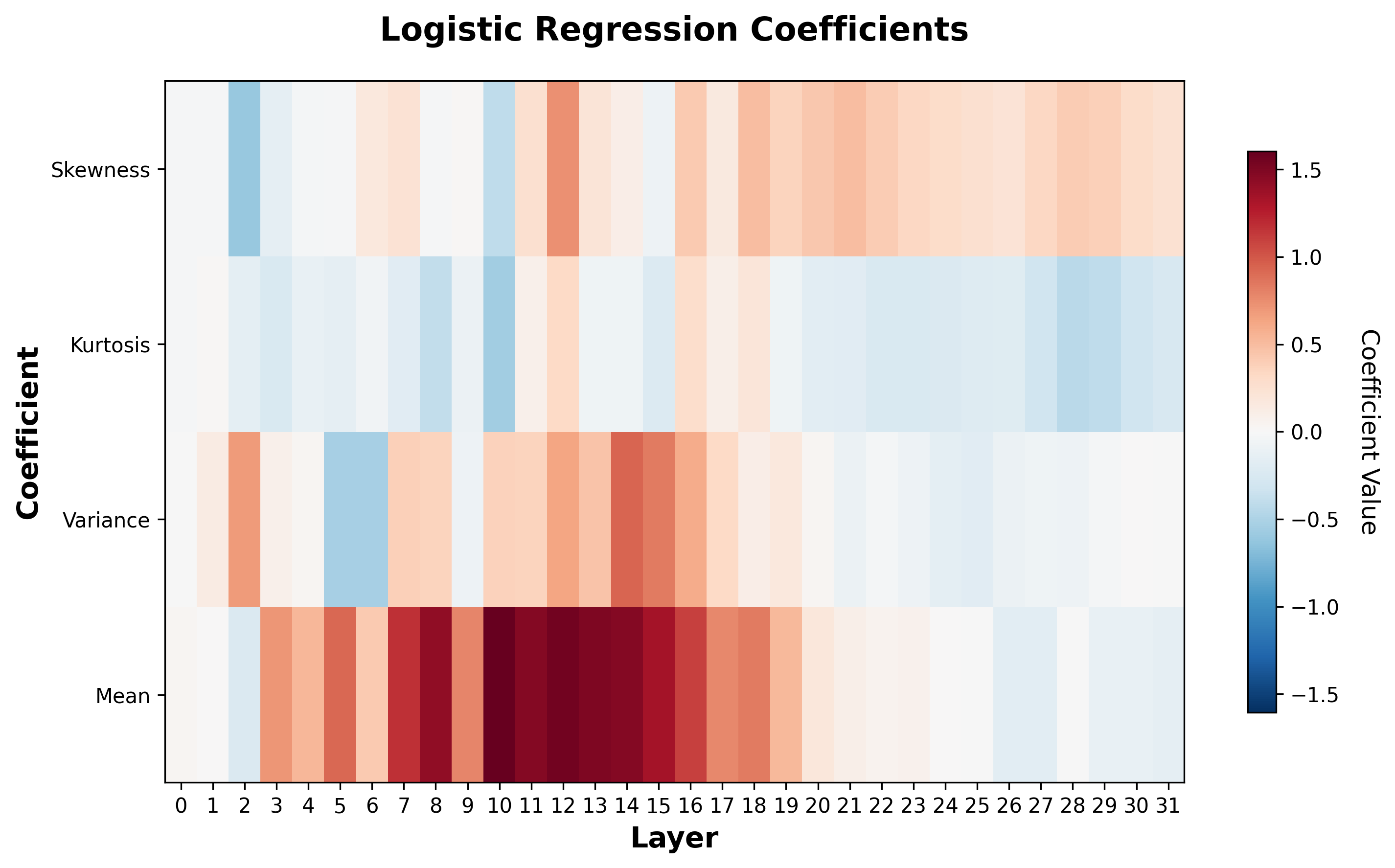}
        \caption{KriKri-Instruct coefficients}
        \label{fig:appendix_coeff_krikri}
    \end{subfigure}
    \hfill
    \begin{subfigure}[t]{0.48\textwidth}
        \centering
        \includegraphics[width=\textwidth]{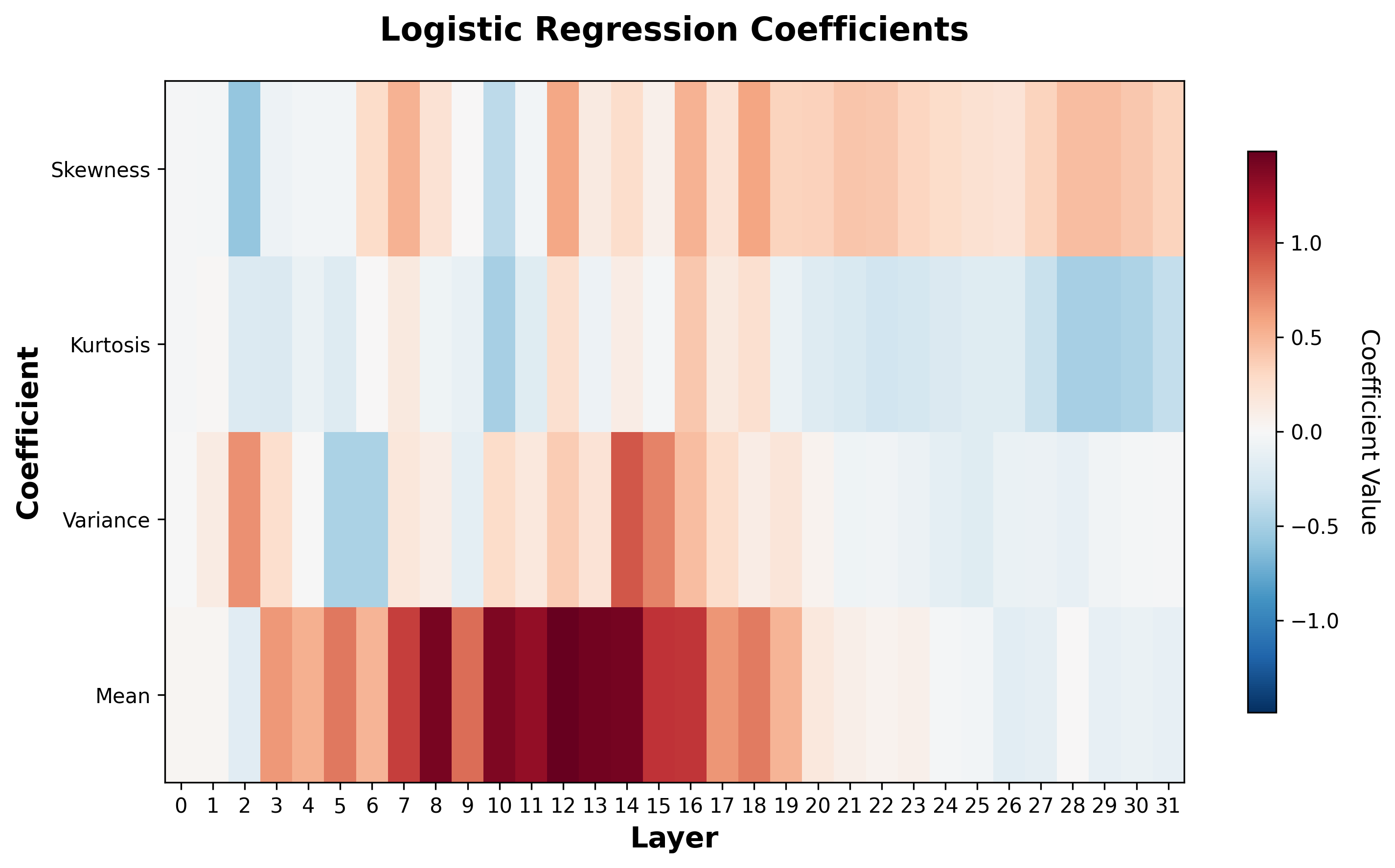}
        \caption{Plutus coefficients}
        \label{fig:appendix_coeff_plutus}
    \end{subfigure}
    \caption{Coefficient plots for the two comparison models on the expanded
    $1024$-sentence stimulus set. Both models show the same broad sign-reversal
    pattern that accompanies the late-to-early below-chance generalization
    regime under the large stimulus set.}
    \label{fig:appendix_coeff_models}
\end{figure*}

To complement the GAT matrices, Figure~\ref{fig:appendix_coeff_models}
displays the corresponding coefficient plots for the same large-set runs. These
show the same broad sign-reversal profile across models: coefficients are
predominantly positive in the middle-to-late layers where decoding is
strongest, and they flip sign in the late regime that drives the below-chance
transfer pattern.

\section{Lexicon and expanded stimulus set}
\label{sec:appendix_lexicon_stimuli}

\subsection{Tokenization \& Forward pass}
\label{sec:tokenization}

All sentences were fed to the model using its native subword tokenizer, which extends Llama~3.1 with Greek-specific units. Inputs were the raw Greek strings from the controlled stimulus set, with sentence-final punctuation preserved and no prompt or few-shot context added. We executed inference-only forward passes with the pretrained weights, disabling caching and dropout, and extracted the complete set of hidden states at every layer and token position. 

The clause-initial complementizer \selectlanguage{greek}\emph{που}\selectlanguage{english} served as a fixed temporal anchor. For each sentence, we partitioned the token sequence into a ``before'' region (up to and including the complementizer) and an ``after'' region. Token–word alignment exploited this anchor and the deterministic SVO/VSO templates: multi-token words (e.g., morphologically complex nouns or verbs) were treated as spans, and their word-level representations were computed as the mean of constituent token vectors. This procedure allowed us to consistently index the determiner and head noun of the matrix subject ($Det~N1$, $N1$), the determiner and head noun inside the relative clause ($Det~N2$, $N2$), and the two verbs (V1 transitive inside the clause, V2 sentence-final). These layer-by-position activations constitute the sole inputs to our probing analyses and all subsequent figures.
\subsection{Lexicon}
\label{sec:appendix_lexicon}

The experimental materials were generated from a fixed lexicon of Greek nouns,
verbs, and determiners. The noun inventory consisted of human-denoting stems
with masculine and feminine forms in both singular and plural. The verbal
inventory included distinct transitive and intransitive paradigms for singular
and plural forms. Tables~\ref{tab:lexicon_details} and
\ref{tab:lexicon_verbs_determiners} report the complete lexicon used by the
stimulus generator.

\begin{table*}[!t]
\centering
\footnotesize
\setlength{\tabcolsep}{5pt}
\renewcommand{\arraystretch}{0.95}
\caption{Noun lexicon used to generate the experimental stimuli.}
\label{tab:lexicon_details}
\begin{tabular}{p{0.18\textwidth}p{0.39\textwidth}p{0.35\textwidth}}
\hline
\textbf{Category} & \textbf{Greek} & \textbf{English translation} \\
\hline
\textbf{Masculine singular} &
\selectlanguage{greek}δάσκαλος, μαθητής, φοιτητής, γυμναστής, καθηγητής,
λογιστής, μεταφραστής, σχεδιαστής, πωλητής, κομμωτής, νοσοκόμος, χορευτής,
αθλητής, μάγειρας, εργάτης, υποψήφιος, ερευνητής, διευθυντής,
προγραμματιστής, σκηνοθέτης, τραγουδιστής, νοσηλευτής, τεχνίτης,
ζαχαροπλάστης, σερβιτόρος\selectlanguage{english} &
teacher, student, university student, gym teacher, professor, accountant,
translator, designer, salesman, hairdresser, nurse, dancer, athlete, chef,
worker, candidate, researcher, director, programmer, director, singer, nurse,
craftsman, pastry chef, waiter \\
\hline
\textbf{Masculine plural} &
\selectlanguage{greek}δάσκαλοι, μαθητές, φοιτητές, γυμναστές, καθηγητές,
λογιστές, μεταφραστές, σχεδιαστές, πωλητές, κομμωτές, νοσοκόμοι, χορευτές,
αθλητές, μάγειρες, εργάτες, υποψήφιοι, ερευνητές, διευθυντές,
προγραμματιστές, σκηνοθέτες, τραγουδιστές, νοσηλευτές, τεχνίτες,
ζαχαροπλάστες, σερβιτόροι\selectlanguage{english} &
teachers, students, university students, gym teachers, professors,
accountants, translators, designers, salesmen, hairdressers, nurses, dancers,
athletes, chefs, workers, candidates, researchers, directors, programmers,
directors, singers, nurses, craftsmen, pastry chefs, waiters \\
\hline
\textbf{Feminine singular} &
\selectlanguage{greek}δασκάλα, μαθήτρια, φοιτήτρια, γυμνάστρια, καθηγήτρια,
λογίστρια, μεταφράστρια, σχεδιάστρια, πωλήτρια, κομμώτρια, νοσοκόμα,
χορεύτρια, αθλήτρια, μαγείρισσα, εργάτρια, υποψήφια, ερευνήτρια,
διευθύντρια, προγραμματίστρια, σκηνοθέτρια, τραγουδίστρια, νοσηλεύτρια,
τεχνίτρια, ζαχαροπλάστρια, σερβιτόρα\selectlanguage{english} &
teacher, student, university student, gym teacher, professor, accountant,
translator, designer, saleswoman, hairdresser, nurse, dancer, athlete, chef,
worker, candidate, researcher, director, programmer, director, singer, nurse,
craftswoman, pastry chef, waitress \\
\hline
\textbf{Feminine plural} &
\selectlanguage{greek}δασκάλες, μαθήτριες, φοιτήτριες, γυμνάστριες,
καθηγήτριες, λογίστριες, μεταφράστριες, σχεδιάστριες, πωλήτριες, κομμώτριες,
νοσοκόμες, χορεύτριες, αθλήτριες, μαγείρισσες, εργάτριες, υποψήφιες,
ερευνήτριες, διευθύντριες, προγραμματίστριες, σκηνοθέτριες,
τραγουδίστριες, νοσηλεύτριες, τεχνίτριες, ζαχαροπλάστριες,
σερβιτόρες\selectlanguage{english} &
teachers, students, university students, gym teachers, professors,
accountants, translators, designers, saleswomen, hairdressers, nurses,
dancers, athletes, chefs, workers, candidates, researchers, directors,
programmers, directors, singers, nurses, craftswomen, pastry chefs,
waitresses \\
\hline
\end{tabular}
\end{table*}

\begin{table*}[!t]
\centering
\footnotesize
\setlength{\tabcolsep}{5pt}
\caption{Verb and determiner lexicon used to generate the experimental
stimuli.}
\label{tab:lexicon_verbs_determiners}
\begin{tabular}{p{0.17\textwidth}p{0.35\textwidth}p{0.39\textwidth}}
\hline
\textbf{Category} & \textbf{Greek} & \textbf{English translation} \\
\hline
\textbf{Intransitive singular} &
\selectlanguage{greek}φεύγει, κλαίει, σκέφτεται, γελάει, βήχει,
πονάει\selectlanguage{english} &
leaves, cries, thinks, laughs, coughs, hurts \\
\hline
\textbf{Intransitive plural} &
\selectlanguage{greek}φεύγουν, κλαίνε, σκέφτονται, γελάνε, βήχουν,
πονάνε\selectlanguage{english} &
leave, cry, think, laugh, cough, hurt \\
\hline
\textbf{Transitive singular} &
\selectlanguage{greek}συμπαθεί, αντιπαθεί, αγαπάει, μισεί, γνωρίζει,
θαυμάζει, εκτιμά\selectlanguage{english} &
likes, dislikes, loves, hates, knows, admires, appreciates \\
\hline
\textbf{Transitive plural} &
\selectlanguage{greek}συμπαθούν, αντιπαθούν, αγαπούν, μισούν, γνωρίζουν,
θαυμάζουν, εκτιμούν\selectlanguage{english} &
like, dislike, love, hate, know, admire, appreciate \\
\hline
\textbf{Determiners} &
\selectlanguage{greek}ο, η, οι &
\selectlanguage{english}the (masculine singular, feminine singular, plural) \\
\hline
\end{tabular}
\end{table*}

\subsection{Expanded stimulus set}
\label{sec:appendix_expanded_stimuli}

In addition to the main 128-sentence experiment, we generated an expanded
stimulus set containing $1024$ sentences for the sensitivity analyses reported
in the appendix. The larger set preserves the same design logic as the main
experiment: it contains the same 32 fully counterbalanced linguistic
conditions, enforces distinct lexical stems for $N1$ and $N2$, and varies only
word order while keeping sentence meaning fixed within each minimal-pair
contrast.

\begin{table}[!t]
\centering
\footnotesize
\caption{Summary of the two generated stimulus sets.}
\label{tab:stimulus_set_summary}
\begin{tabular}{lcc}
\hline
\textbf{Set} & \textbf{Sentences} & \textbf{Items per condition cell} \\
\hline
Main experiment & 128 & 4 \\
Expanded appendix set & $1024$ & 32 \\
\hline
\end{tabular}
\end{table}

The expanded set was used to test whether the decoding and generalization
effects reported in the main text remain stable when the number of lexicalized
sentence instances per condition is substantially increased. The qualitative
pattern is preserved under this larger sampling regime, indicating that the
reported effects are not an artifact of the smaller 128-item stimulus set. 
The full sentence lists are available in the \href{https://osf.io/5d3w8/overview?view_only=d42af279745543808cc377b9f96cb1af}{OSF project} under \texttt{code/stimuli}
(\texttt{greek\_sentences\_128.csv} and \texttt{greek\_sentences\_1024.csv}); see \url{https://osf.io/5d3w8/overview?view_only=d42af279745543808cc377b9f96cb1af}.

\clearpage
\subsection{Example generated sentences}
\label{sec:appendix_examples}

Below we provide three representative minimal pairs from the 128-sentence set,
illustrating how the template produces matched SVO and VSO stimuli. In each
pair, only the word order inside the bracketed relative clause differs; all
lexical items, morphological forms, and propositional content are identical.

\vspace{6pt}

\noindent\textbf{Pair 1 (masculine singular, transitive = \selectlanguage{greek}συμπαθεί\selectlanguage{english}):}

\vspace{3pt}
\noindent\textit{SVO:}\ \
\selectlanguage{greek}\textit{Ο δάσκαλος που [η μαθήτρια συμπαθεί] φεύγει.}\selectlanguage{english}\\
\hspace*{1.4em}\textsc{def.m.nom} teacher.\textsc{nom} \textsc{comp} [\textsc{def.f.nom} student.\textsc{nom} like.\textsc{3sg}] leave.\textsc{3sg}\\
\hspace*{1.4em}`The teacher that the student likes is leaving.'

\vspace{3pt}
\noindent\textit{VSO:}\ \
\selectlanguage{greek}\textit{Ο δάσκαλος που [συμπαθεί η μαθήτρια] φεύγει.}\selectlanguage{english}\\
\hspace*{1.4em}\textsc{def.m.nom} teacher.\textsc{nom} \textsc{comp} [like.\textsc{3sg} \textsc{def.f.nom} student.\textsc{nom}] leave.\textsc{3sg}\\
\hspace*{1.4em}`The teacher that the student likes is leaving.'

\vspace{8pt}

\noindent\textbf{Pair 2 (feminine singular, transitive = \selectlanguage{greek}θαυμάζει\selectlanguage{english}):}

\vspace{3pt}
\noindent\textit{SVO:}\ \
\selectlanguage{greek}\textit{Η νοσοκόμα που [ο ερευνητής θαυμάζει] γελάει.}\selectlanguage{english}\\
\hspace*{1.4em}\textsc{def.f.nom} nurse.\textsc{nom} \textsc{comp} [\textsc{def.m.nom} researcher.\textsc{nom} admire.\textsc{3sg}] laugh.\textsc{3sg}\\
\hspace*{1.4em}`The nurse that the researcher admires is laughing.'

\vspace{3pt}
\noindent\textit{VSO:}\ \
\selectlanguage{greek}\textit{Η νοσοκόμα που [θαυμάζει ο ερευνητής] γελάει.}\selectlanguage{english}\\
\hspace*{1.4em}\textsc{def.f.nom} nurse.\textsc{nom} \textsc{comp} [admire.\textsc{3sg} \textsc{def.m.nom} researcher.\textsc{nom}] laugh.\textsc{3sg}\\
\hspace*{1.4em}`The nurse that the researcher admires is laughing.'

\vspace{8pt}

\noindent\textbf{Pair 3 (masculine plural, transitive = \selectlanguage{greek}μισούν\selectlanguage{english}):}

\vspace{3pt}
\noindent\textit{SVO:}\ \
\selectlanguage{greek}\textit{Οι αθλητές που [οι σχεδιαστές μισούν] κλαίνε.}\selectlanguage{english}\\
\hspace*{1.4em}\textsc{def.pl.nom} athlete.\textsc{nom.pl} \textsc{comp} [\textsc{def.pl.nom} designer.\textsc{nom.pl} hate.\textsc{3pl}] cry.\textsc{3pl}\\
\hspace*{1.4em}`The athletes that the designers hate are crying.'

\vspace{3pt}
\noindent\textit{VSO:}\ \
\selectlanguage{greek}\textit{Οι αθλητές που [μισούν οι σχεδιαστές] κλαίνε.}\selectlanguage{english}\\
\hspace*{1.4em}\textsc{def.pl.nom} athlete.\textsc{nom.pl} \textsc{comp} [hate.\textsc{3pl} \textsc{def.pl.nom} designer.\textsc{nom.pl}] cry.\textsc{3pl}\\
\hspace*{1.4em}`The athletes that the designers hate are crying.'

\end{document}